\documentclass{article}
\usepackage{graphicx} 
\usepackage{booktabs}
\usepackage{amsmath}
\usepackage{url}
\usepackage{tablefootnote}
\usepackage{authblk}

\title{JAI-1: A Thai-Centric Large Language Model}

\author[1,2]{Attapol~T.~Rutherford}
\author[1]{Jullajak~Karnjanaekarin}
\author[1]{Narongkorn~Panitsrisit}
\author[1]{Pontakorn~Trakuekul}
\author[1]{Sumana~Sumanakul}
\author[1]{Natchanon Pollertlam}
\author[3]{Upstage~Team}
\affil[1]{Jasmine Technology Solution, Thailand} 
\affil[2]{Department of Linguistics, Chulalongkorn University, Thailand} 
\affil[3]{Upstage, South Korea}
\affil[ ]{\texttt{jts.ai.team@gmail.com}}

\date{}

\begin{document}

\maketitle

\begin{abstract}

This technical report introduces JAI-1, a Thai-centric language model with 75B parameters. Recent Thai models have primarily relied on existing open-source models, applying additional training without structural modifications to specialize in Thai. However, this approach risks eroding pre-existing knowledge in the model’s parameter space during the injection of Thai-specific information, as optimized parameters for general tasks may conflict with new linguistic requirements. In contrast, JAI-1 adopts a upscaling strategy: starting from a smaller, high-performing English open-source LLM, we expanded its parameter space and utilized the newly allocated capacity to systematically integrate Thai-language knowledge. This methodology not only preserves the original model’s general intelligence but also establishes a unique architecture distinct from other open-source models, enabling scalable future enhancements.
 
During pre-training, JAI-1 was exposed to 1.5T tokens, including over 300B Thai language tokens. This was followed by post-training stages—supervised fine-tuning and alignment tuning—using more than 600K instruction-based examples. The final model demonstrated superior performance compared to Typhoon2-70B on Thai-centric benchmarks (IFEval-TH, MT-Bench-TH, and JAI-Hall-Bench), validating the efficacy of its upscaling and knowledge-integration framework.

\end{abstract}

\newpage
\tableofcontents
\newpage

\section{Introduction}

Large Language Models (LLMs) have rapidly become ubiquitous in daily life, transforming how individuals access, organize, and leverage information to address everyday challenges. These models synthesize vast global knowledge, offering profound assistance in human activities. However, their continued advancement hinges on two critical factors: a deep understanding of linguistic and cultural contexts across diverse regions~\cite{adilazuarda2024towards}, and the development of resource-efficient architectures tailored to dynamic, real-world applications~\cite{bai2024beyond}.

Despite the existence of over 7000+ languages worldwide, the majority of training data remains concentrated in English, creating significant disparities in representation~\cite{yang2024language}. This imbalance is particularly evident in Thailand, where linguistic and cultural nuances are underrepresented in global datasets~\cite{limkonchotiwat2025assessing}. Consequently, existing LLMs often struggle to capture the unique characteristics of Thai language and culture~\cite{nguyen2024seallms}, limiting their effectiveness in localized contexts.

To bridge this gap, Thai-centric LLMs such as Typhoon~\cite{typhoon,typhoon2} and OpenThai-GPT \cite{openthaigpt15} have recently emerged. Typhoon1 (7B)~\cite{typhoon} adapted Mistral-7B through additional pre-training on a Thai corpus, evolving into Typhoon 1.5~\cite{typhoon15} built on Qwen 1.5~\cite{qwen}, and the latest Typhoon 2~\cite{typhoon2}, leveraging Llama 3.1 with Thai data injection. Similarly, OpenThaiGPT~\cite{openthaigpt15} began by fine-tuning Llama 2, and its recent v1.5 series utilizes Qwen 2.5 checkpoints tuned on multi-million Thai instruction pairs. These Thai-centric LLMs follow an open-source paradigm: injecting proprietary Thai data into advancing backbones—an efficient community-driven strategy. However, this model construction risks eroding pre-existing knowledge in the model’s parameter space—catastrophic forgetting—when specializing for Thai~\cite{catastrophic_forgetting,catastrophicforgettingllms}. Additionally, lack of visibility into upstream architectures and data constrains transparency, auditing capacity, and bias control.

Building upon this momentum, our project also develops a Thai-centric LLM, named JAI-1(75B), designed with two considerations in mind. First, we utilize trillion-level pretraining data: a mix of English and Thai corpora, where English maintains and enhances general capabilities while Thai injects localized knowledge. Unlike previous models, our dataset exceeds 1.5T tokens for pre-training phase. Second, we redesign the model architecture: rather than finetuning an English-dominant model (which risks overwriting existing knowledge), we upscale a smaller open model to 75B parameters and define a new architecture, providing dedicated capacity for Thai specialization. Through these two innovations, we progressively decouple from upstream open-source backbones and gain full control over cultural and knowledge biases present in the final model. 

Extensive evaluations demonstrate JAI-1’s superiority in Thai-language benchmarks, including Thai-MT-Bench and Thai-IFEval, alongside robust performance in cultural understanding tasks. By bridging the representation gap and optimizing for localized utility, this work advances the development of culturally aware, efficient LLMs for underrepresented linguistic communities.
\section{JAI-1 Overview}

\subsection{Upscaling Strategy for Architecture Design}

The JAI-1 project aims to develop a Thai-language-specific LLM capable of deep understanding of Thai language and cultural contexts through extensive training on Thai datasets. Central to this endeavor is the principle that foundational intellectual capabilities—such as general knowledge, reasoning, and multitask proficiency—must be established before enhancing Thai-specific competencies. To achieve this, we adopt an up-scaling strategy: starting with a smaller, knowledge-rich base model, we expand its architecture to integrate additional linguistic and cultural knowledge, thereby creating a high-performing Thai-centric LLM. This approach bypasses the inefficiencies of training from scratch by leveraging pre-trained open-source models as a robust starting point~\cite{lottery,solar-dus}.

\begin{figure}[h!]
    \centering
    \includegraphics[width=0.9\linewidth]{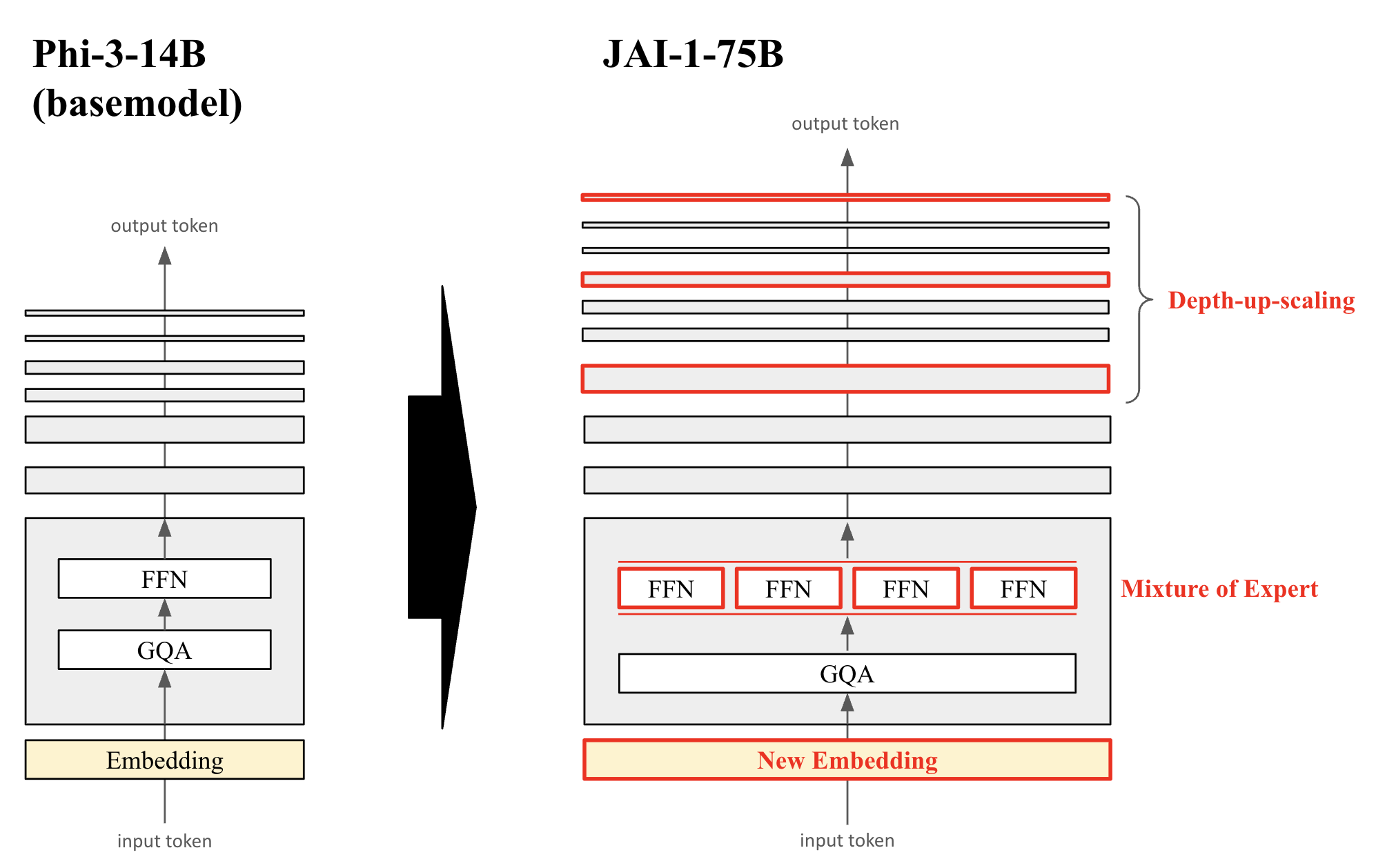}
    \caption{Architectural overview of the JAI-1 (75B) model. JAI-1 is a Thai-optimized LLM developed by scaling the Phi-3-medium (14B) architecture. While Phi-3-medium excels in general knowledge benchmarks, JAI-1 significantly enhances its parameter count to 75B through depth-up-scaling (DUS) and mixture-of-experts (MoE) techniques. Additionally, the model incorporates a redesigned Thai-optimized tokenizer to improve representation of Thai script and linguistic features. In the figure, GQA and FFN represent grouped-query attention and feed-forward network layers respectively.}
    \label{fig:upscaling-strategy}
\end{figure}

To comply with licensing requirements, we evaluated open-source models with permissive terms. In April 2024, Microsoft’s phi-3-medium (14B)~\cite{phi3} emerged as a compelling candidate. Despite its compact architecture, it exhibited robust foundational capabilities, achieving a MMLU score of 78.0—notably competitive with larger models like Llama-3-70B-Instruct (80.2) and GPT-4 Turbo (84.0 While these larger models prioritize raw performance, they lacked the efficiency and adaptability required for our Thai-focused use case. The phi-3-medium’s optimal balance of size and performance made it a strategic choice for scaling our system, particularly for deployment in resource-constrained environments.

To scale phi-3-medium to 75B parameters—over five times increase—we implemented three core strategies:
\begin{itemize}
    \item \textbf{Tokenizer Adaptation}: The model’s tokenizer was redesigned to optimize processing of Thai script, addressing linguistic nuances such as tonal markers and compound words
    \item \textbf{Depth-Up-Scaling (DUS)~\cite{solar-dus}}: The architecture was expanded through additional transformer layers and parameter increases, enhancing its capacity for complex reasoning and knowledge retention.
    \item \textbf{Mixture-of-Experts (MoE)~\cite{moe-survey}}: MoE layers were integrated to enable dynamic task specialization, allowing subsets of the model to focus on Thai-specific challenges like idiomatic expressions and cultural references.
\end{itemize}
This strategic expansion preserved the base model’s foundational knowledge while introducing Thai-centric enhancements. The upscaling process was visualized in Figure~\ref{fig:upscaling-strategy}, which outlines architectural modifications. 

\subsection{Training Pipeline}

Following the architectural modifications, the JAI-1 model underwent a standard large language model (LLM) training pipeline comprising pre-training and post-training phases. The pre-training phase focused on equipping the model with general linguistic and reasoning capabilities through next token prediction on large-scale raw Thai-language datasets. This stage enabled the model to acquire core knowledge about syntax, semantics, and world phenomena, forming the bedrock for subsequent task-specific refinements. The post-training phase was divided into two distinct sub-stages to align the model with practical use cases:
\begin{itemize}
    \item \textbf{Supervised Finetuning (SFT)}: The model was trained on high-quality datasets formatted with chat templates, teaching it to follow instructions and generate contextually appropriate responses.
    \item \textbf{Alignment Tuning}: This stage introduced learning from human feedback to refine the model’s behavior. 
\end{itemize}
By structuring post-training into SFT and alignment phases, JAI-1 achieved both task proficiency and value-aligned behavior, critical for deployment in real-world applications. The following sessions will provide more detailed explanations for pre-training, supervised finetuning and alignment tuning.
\section{JAI-1 Architecture}

As described in Section 2.1, we applied three upscaling strategies: tokenizer adaptation, DUS and MoE. This section describes each upscaling approach and the finally identified model configuration.

\subsection{Upscaling Strategies}

Publicly available pre-trained language models (PLMs) are highly optimized for specific data distributions, which often leads to challenges when performing additional continual pre-training (CPT). These challenges include: (1) difficulties in smoothly integrating new knowledge~\cite{difficulty_cpt} and (2) catastrophic forgetting~\cite{catastrophicforgettingllms}, where previously acquired knowledge is lost. Since we plan to inject a large volume of Thai language data, we employ an upscaling strategy to ensure the model parameters can accommodate additional Thai linguistic information and knowledge. Specifically, we implement three key approaches: Tokenizer adaptation, DUS and MoE. This subsection details each of these methods.

\subsubsection{Tokenizer Adaptation}

Phi-3-medium employs a character-level Byte Pair Encoding (BPE) tokenizer with a vocabulary size of 32,064 tokens, which is primarily optimized for English subwords. While the model supports multilingual processing, its original vocabulary lacks sufficient coverage for Thai-specific linguistic patterns. To address this limitation, we expanded the tokenizer by incorporating Thai-language corpora into the BPE training process. This extension increased the total vocabulary size to 64,000 tokens, enhancing the model’s ability to represent Thai character sequences effectively. Detailed principles of BPE tokenization are referenced in \cite{huggingface2025bpe}.

The expansion of the vocabulary introduced a critical challenge: the original 32,064 token embeddings in phi-3-medium could not accommodate the additional 31,936 tokens. Randomly initializing these new embeddings risks destabilizing training by introducing noisy parameters and increasing computational costs. To mitigate this, we propose a token embedding merging strategy that leverages semantic relationships between new and existing tokens. The method operates in two steps:
\begin{itemize}
    \item \textbf{Origin Token Matching}: Each new token is decomposed into constituent subwords using the original tokenizer, yielding a set of existing (origin) tokens.
    \item \textbf{Embedding Aggregation}: The embeddings of these origin tokens are averaged to initialize the new token’s embedding, enabling semantically grounded initialization rather than random assignment.
\end{itemize}
This approach preserves semantic consistency for 64,000 tokens, while only a negligible fraction (e.g., tokens with no matching origin subwords) required random initialization. By reducing reliance on arbitrary parameter values, this method enhances training stability and efficiency when adapting the model to new linguistic domains.

To quantify the impact of our adaptation on Thai‑text token efficiency, we measured the total token count and average tokens per character (TPC) using the output of the training split from the WangchanThaiInstruct corpus\footnote{\url{https://huggingface.co/datasets/airesearch/WangchanThaiInstruct}}. Table~\ref{tab:token_efficiency} reports the results for our adapted tokenizer (JAI), the original Phi‑3, and other state‑of‑the‑art multilingual baselines.

\begin{table}[h]
  \centering
  \begin{tabular}{lrr}
    \toprule
    \textbf{Tokenizer}     & \textbf{Total Tokens} & \textbf{TPC}  \\
    \midrule
    JAI                    & \textbf{13,436,003}            & \textbf{0.2913} \\
    phi‑3                  & 47,562,113            & 1.0311         \\
    phi‑4                  & 18,914,748            & 0.4100         \\
    Qwen‑3                 & 25,328,006            & 0.5491         \\
    Gemma‑3                & 15,763,728            & 0.3417         \\
    Llama‑3                & 21,454,143            & 0.4651         \\
    \bottomrule
  \end{tabular}
  \caption{Token efficiency on Thai text: lower TPC is better.}
  \label{tab:token_efficiency}
\end{table}

As shown, our adapted JAI tokenizer achieves a TPC of 0.2913, over a 71\% reduction compared to the original Phi‑3, and outperforms all multilingual baselines in compactness. This demonstrates that incorporating Thai‑specific BPE rules yields a substantially more efficient token representation.

\subsubsection{Advanced Depth-Up-Scaling}

\begin{figure}[h!]
    \centering
    \includegraphics[width=0.8\linewidth]{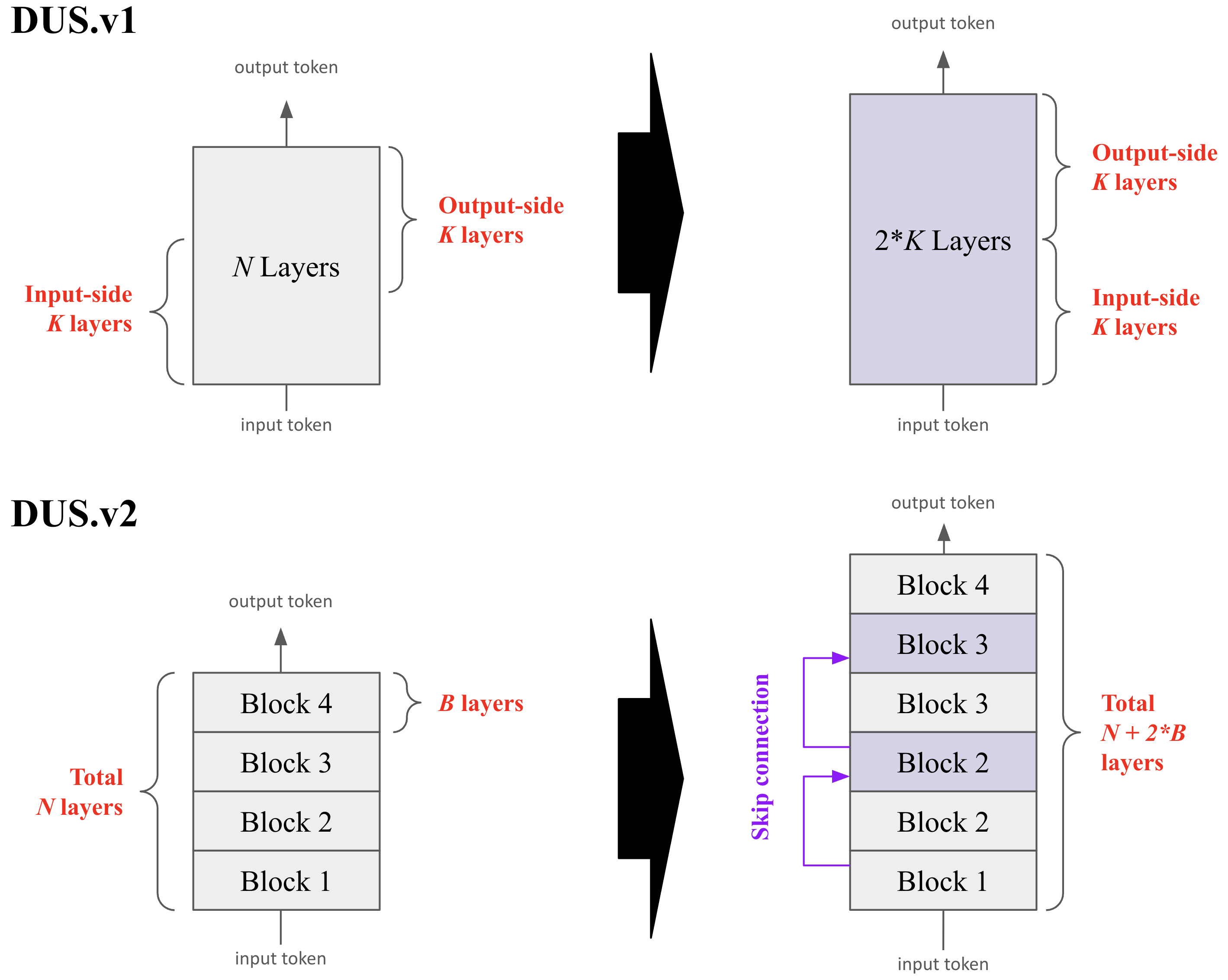}
    \caption{Visual illustration of the key differences between two depth-scaling strategies in the DUS framework. The original DUS method, referred to as DUS.v1, extends model depth by selecting K layers from the input side and K layers from the output side, then concatenating these two sets to form a deeper architecture with 2K layers. In contrast, DUS.v2 introduces a block-level approach: the original model is divided into multiple blocks, each containing B consecutive layers, and depth is scaled by duplicating selected blocks in their original sequential order.}
    \label{fig:dus-duplication}
\end{figure}

DUS is a technique that increases model depth by duplicating existing layers, reducing the computational cost of training from scratch. The original DUS method merges K layers from the input side and K layers from output side to build a deeper network with 2*K layers. In this work, we applied standard DUS to extend a 40-layer phi-3-medium model to 64 layers by setting K=32. However, unlike with Llama-based models~\cite{solar-dus}, we observed significant instability in the early training phase when applying this method to the phi-3-medium architecture. 

Since depth-scaling via DUS can be achieved through various strategies beyond the original method, we experimented with multiple configurations and identified two key structural improvements that led to DUS.v2: Block-level Duplication and Skip Connections Between Identical Blocks.
\begin{itemize}
    \item \textbf{Block-level Duplication:} Unlike standard DUS, which duplicates individual layers, our approach organizes the entire model into blocks and replicates selected blocks in their original positions.
    \item \textbf{Skip Connections Between Identical Blocks:} In the original model, information flows through blocks in a specific sequence. Block-level Duplication disrupts this flow by repeating blocks, which can hinder training efficiency and performance. To mitigate this, we introduced additional skip connections between duplicated blocks. These connections preserve the original information-processing pathways, allowing the model to leverage learned representations more effectively.
\end{itemize}

Figure~\ref{fig:dus-duplication} provides a visual explanation of these two duplication strategies of DUS.v1 and DUS.v2. The most significant distinction between the two methods lies in how they preserve the layer order of the base model. In DUS.v1, the two defined blocks (input-side and output-side) retain the original layer sequence of the base model. The only disruption to the base model’s layer order occurs at the single merging point where the two blocks are concatenated, which explains why DUS.v1 often leads to suboptimal performance without additional fine-tuning. While this simplistic block definition and merging strategy is straightforward, it lacks flexibility to accommodate diverse DUS strategies across different model architectures.

In contrast, DUS.v2 first divides the base model’s layers into fixed-size blocks, then selectively duplicates chosen blocks to scale depth. Though this approach offers greater flexibility for depth scaling, it introduces more disruptions to the original layer sequence compared to DUS.v1. For instance, in Figure~\ref{fig:dus-duplication}, the repetition of Block 2 and Block 3 creates points where the original information flow is altered, potentially affecting model performance.

To address disruptions in the base model’s information flow caused by block duplication, DUS.v2 incorporates additional skip connections. As shown in Figure 2, when a block is duplicated (e.g., Block 2 or Block 3 in purple), a skip connection is established from the previous block in the original sequence. This can be expressed mathematically as:
\begin{equation}
    \text{Output}_{n, (d)} = \text{Block}_{n}( x + \alpha_{n, (d)} * \text{Output}_{n-1, (\max d)})
\end{equation}
where $n$ indicates the block index of the base model (origin) and $d$ represents the duplication index of the DUS model. $\text{Output}_{n, (d)}$ is the output of the $d$-th duplicated block of $\text{Block}_{n}$. For examples, the output of the first Block 2 in grey is represented as $\text{Output}_{2, (1)}$ and the output of the second Block 2 in purple is denoted as $\text{Output}_{2, (2)}$. $x$ is the input feature from the previous layer in the DUS model and $\alpha_{n, (d)}$ is a scaling factor for the original information process. $\max d$ indicates the latest duplication index. Importantly, $\alpha_{n, (1)}=0$ for the first block in each duplication sequence, as no skip connection is needed before duplication begins. 

This approach preserves the base model’s inherent information flow within the duplicated blocks. For example, in Figure 2, the duplicated Block 2 and Block 3 receive skip connections from the original Block 1 and Block 2, respectively. This mimics the base model’s original flow (Block 1 → Block 2 → Block 3), ensuring continuity even after depth scaling. Internal evaluations confirmed that this modification significantly improves training stability and learning efficiency post-DUS, contributing to more consistent performance gains.

\begin{table}[h!]
    \centering
    \begin{tabular}{ l|l|c|c}
    \toprule
    blocks & origin layers & block usage & \# of layers after DUS.v2 \\
    \midrule
    Block-1 & layer1-4 & 1 & 4 \\
    Block-2 & layer5-8 & 1 & 4 \\
    Block-3 & layer9-12 & 1 & 4 \\
    Block-4 & layer13-16 & \textbf{2} & \textbf{8} \\
    Block-5 & layer17-20 & \textbf{3} & \textbf{12} \\
    Block-6 & layer21-24 & \textbf{3} & \textbf{12} \\
    Block-7 & layer25-28 & \textbf{2} & \textbf{8} \\
    Block-8 & layer29-32 & 1 & 4 \\
    Block-9 & layer31-36 & 1 & 4 \\
    Block-10 & layer37-40 & 1 & 4 \\
    \bottomrule
    \end{tabular}
    \caption{Block-level duplication applied to JAI-1. Block1-3 and Block 8-10 are not duplicated in the DUS process. The middle blocks are duplicated for 1 or 2 times. Finally, the number of layers of the model after DUS.v2 becomes 64.}
    \label{tab:block-level-duplication}
\end{table}

In JAI-1, we empirically determined that a block size of 4 layers was optimal. We excluded some input-side and output-side blocks from replication to scale the model’s depth while maintaining structural integrity. Table~\ref{tab:block-level-duplication} illustrates which blocks are replicated and how many times.
 
\subsubsection{Mixture of Expert Design}

Motivated by advancements in models like Mistral-8x22B-v0.1 and Phi-3.5-MoE, we incorporate a basic MoE framework into JAI-1. The MoE architecture is structured as follows: (1) At each transformer layer, M=4 experts (feed-forward networks, FFNs) are defined and (2) A routing layer selects K=2 active experts before processing inputs through the FFN. The outputs from the selected experts are then aggregated. For a comprehensive overview of MoE principles, refer to \cite{moe-survey}.

A key challenge in MoE initialization arises when one expert is initialized with pretrained FFN weights (from DUS.v2), while the remaining three are randomly initialized. This asymmetry can introduce routing bias, where the model disproportionately favors the pretrained expert. To mitigate this, we employ a heuristic strategy inspired by weight-space exploration:
\begin{itemize}
    \item Pretraining and Snapshot Collection: We first train the non-MoE version of JAI-1 for 6K steps. During this phase, we save 3 snapshots of the FFN weights at intervals of 2K steps (e.g., steps 2K, 4K, 6K).
    \item Expert Initialization: The initial (step 0) FFN weights are assigned to Expert 0. The 3 snapshots are used to initialize Experts 1–3, avoiding random initialization.
\end{itemize}
By leveraging learned weights from different training stages, we reduce routing bias and ensure balanced expert selection. This approach avoids introducing random noise into the weight space while identifying viable expert candidates.

\subsection{Architecture Configuration}

Here, we introduce the final JAI-1 architectural configuration after applying three upscaling methods. Since expanding phi-3-medium, JAI-1 retains its origin layer design: embedding size, intermediate size, activation method, normalization method and the number of attention heads. However, modifications from the tokenizer adaptation, DUS and MoE, increase vocabulary size, the number of layers and the number of experts, collectively achieving a parameter scale-up of over 5x. Table~\ref{tab:configureation} presents the model architecture configurations of phi-3-medium and JAI-1. As shown, while most design options are shared, differences exist in terms of model size.

\begin{table}[h!]
    \centering
    \begin{tabular}{ll|c|c}
    \toprule
         & Attribute & phi-3-medium & JAI-1  \\
         \midrule
    Model size & Parameter size & 14.0B & \textbf{75.3B}  \\
         & Embedding size & 5120 & 5120 \\
         & Intermediate size & 17920 & 17920 \\
         & \# of experts & 1 & \textbf{4} \\
         & \# of layers & 40 & \textbf{64} \\
         & Vocabulary size & 32064 & \textbf{64000} \\
         \midrule
    Layer design & Positional encoding & RoPE($\theta$=10K)~\cite{rope} & RoPE($\theta$=\textbf{1M})~\cite{rope} \\
         & Sliding window & 2047 & \textbf{Null} \\
         & Activation & SiLU~\cite{sliu} & SiLU~\cite{sliu} \\
         & Normalization & RMSNorm~\cite{rmsnorm} & RMSNorm~\cite{rmsnorm} \\ 
         & Attention layer & GQA~\cite{gqa} & GQA~\cite{gqa} \\
         & \# of heads & 40 & 40 \\
    \bottomrule
    \end{tabular}
    \caption{Architecture configuration of JAI-1. JAI-1 adopts an up-scaling strategy to expand the phi-3-medium base model, increasing vocabulary size, layers, and experts to achieve a total parameter count of 75.3B. While retaining core architectural designs from phi-3-medium, key modifications include adjustments to the sliding window mechanism and RoPE parameters to enhance long-context handling and future token length scalability.}
    \label{tab:configureation}
\end{table}

During the model scaling process, we additionally decided to eliminate the sliding window constraint inherent to phi-3-medium. While a fixed 2K sliding window can improve memory efficiency during serving, it poses challenges in achieving robust performance for long contexts. We identified this limitation in phi-3-medium and, during the scale-up, removed this constraint. Instead, we increased the RoPE~\cite{rope} parameter $\theta$ to adopt a positional encoding method better suited for long-context scenarios.
\section{Pre-training}

The primary challenges in pre-training involve (1) collecting and curating a large-scale Thai language dataset and (2) integrating Thai data without compromising the model’s performance. This section provides an overview of the corpora utilized for JAI-1 training, details the pre-training strategies applied, and presents the outcomes of the pre-training process.

\subsection{Thai Corpora for pre-training}

In this project, we constructed a Thai corpus comprising over 300 billion (B) tokens for pre-training. The corpus is categorized into four distinct groups based on its composition method:
\begin{itemize}
    \item \textbf{Public and open data}: We collected publicly available, pre-processed Thai language data, totaling 61B tokens. During this process, datasets deemed to have excessively low quality were excluded. Statistics and sources of the collected data are detailed in Table~\ref{tab:public_thai_data}.
    \item \textbf{Private data}: This dataset was provided by our partners. It consists of 4B billion tokens that are collected from internet sources that are not publicly available or free.
    \item \textbf{Web-crawled data}: Collected manually using a crawling engine, this dataset required site-specific parsers developed individually for each site. Although high-priority data was prioritized, the total volume remained relatively small at 0.5B tokens. Due to the scale limitations, additional collection was paused after initial development. Statistics and sources are detailed in Table~\ref{tab:crawled_data}.
    \item \textbf{Upstage’s processed data}: This category includes two subsets: translated high-quality data and LP-pipeline-based Common Crawl data. The former leverages Upstage’s proprietary data and translation methods. The latter utilizes Upstage's LP data pipeline~\cite{lp_pipeline} to process a large-scaled Common Crawl dump data. Both underwent deduplication and rule-based filtering. The total size is 239.3B tokens, with statistics and sources listed in Table~\ref{tab:additional_data}
\end{itemize}

In sum, the pre-training dataset consists of 165 + 174 = 340B organic tokens and 65B synthetic tokens, totaling 405B tokens. 

\begin{table}[h!]
    \centering
    \begin{tabular}{l|c|l}
        \toprule
        data name & tokens & source \\
        \midrule
        thaigov-corpus & 0.06B & Thai government website since 2018.02\tablefootnote{\url{https://github.com/PyThaiNLP/thaigov-corpus}} \\
        th-law & 0.10B & Office of the council of state, Thailand\tablefootnote{\url{https://github.com/PyThaiNLP/thai-law}} \\
        th-tcc & 0.00B & Office of the council of state since 1932\tablefootnote{\url{https://github.com/PyThaiNLP/Thai-constitution-corpus}}\\
        th-mC4 & 12.53B & cleaned version of Common Crawl's corpus\tablefootnote{\url{https://huggingface.co/datasets/legacy-datasets/mc4}} \\
        th-oscar & 11.18B & web-based multilingual resources and datasets\tablefootnote{\url{https://huggingface.co/datasets/oscar-corpus/OSCAR-2301}}\\
        th-xlsum & 0.01B & article-summary pairs from BBC\tablefootnote{\url{https://huggingface.co/datasets/csebuetnlp/xlsum}}\\
        th-sum & 0.21B & Thai online news \tablefootnote{\url{https://huggingface.co/datasets/thaisum}} \\
        th-culturaX & 17.36B & culturaX dataset\tablefootnote{\url{https://huggingface.co/datasets/uonlp/CulturaX}}\\
        th-wiki & 0.13B & Thai Wikipedia\tablefootnote{\url{https://huggingface.co/datasets/pythainlp/thai-wiki-dataset-v3/viewer}}\\
        th-TLC & 2.36B & Vajirayana Digital Library\tablefootnote{\url{https://attapol.github.io/tlc.html}} \\
        th-HPLT-v1.2 & 17.17B & monolingual and bilingual corpus\tablefootnote{\url{https://hplt-project.org/datasets/v1.2}} \\
        \bottomrule
    \end{tabular}
    \caption{Public and open Thai corpus used in JAI-1. The total number of tokens from this data is 61B.}
    \label{tab:public_thai_data}
\end{table}

\begin{table}[h!]
    \centering
    \begin{tabular}{l|c|l}
         \toprule
        data name & tokens & source \\
        \midrule
        th-bangkokbiz & 0.104B & \url{https://bangkokbiznews.com} \\
        th-rath & 0.358B & \url{https://thairath.co.th} \\
        th-blognone & 0.002B & \url{https://blognone.com} \\
        th-deckd & 0.002B & \url{https://deck-d.com} \\
        th-mhesi & 0.003B & \url{http://mhesi.go.th} \\
        \bottomrule
    \end{tabular}
    \caption{Crawled data from this project. The total number tokens is about 0.5B. We listed up more crawling sites but found that this was not scalable. We decided to turn to translation of our own corpus and filtering process from a large scaled open-sourced data form Common Crawl.}
    \label{tab:crawled_data}
\end{table}

\begin{table}[h!]
    \centering
    \begin{tabular}{l|c|l}
        \toprule
        data name & tokens & source \\
        \midrule
        th-upstage-translation & 65.30B & Translate Upstage's corpus into Thai. \\
        th-lp-pipline & 174.00B & LP pipeline from Common Crawl~\cite{lp_pipeline} \\
        \bottomrule
    \end{tabular}
    \caption{Private Thai corpus developed in the project. The total number tokens is about 239.3B.}
    \label{tab:additional_data}
\end{table}

\subsection{Pre-training Strategy}

The JAI-1 architecture is an upscaled version of a 14B model, requiring dual objectives during training: recovering performance lost during upscaling while enhancing baseline capabilities, and integrating Thai language data. To achieve this, we divided the training into three distinct phases: (1) English-centric pre-training, (2) Thai data injection, and (3) context window expansion. 

\begin{itemize}
    \item \textbf{Phase 1. English-Centric Pre-training}: This phase focused on restoring performance degraded during upscaling and further strengthening baseline capabilities. A key challenge was curating English datasets to maintain or recover the base model’s performance level of 78 points (MMLU score). The base model was highly optimized on proprietary, high-quality datasets that were not publicly available. After upscaling, the initial MMLU score dropped to 62 points due to architectural changes, necessitating the development of a tailored data recipe to restore it to the original 78 points.
    \item \textbf{Phase 2. Thai Data Injection}: In this phase, we integrated the optimized data recipe from Phase 1 with Thai language datasets (defined in Section 4.1) to prevent catastrophic forgetting. The training data distribution was balanced to allocate 27\% of the total dataset to Thai corpus, with the remaining 73\% consisting of multilingual data (primarily English) to preserve the English performance gains from Phase 1.
    \item \textbf{Phase 3. Context Window Expansion}: The base model natively supported a 4K context window, which was preserved during Phases 1 and 2 for performance stability. In the final phase, we expanded the context window to 32K using the sequence-packing method~\cite{huggingface_packing} on the existing data recipe.
\end{itemize}

The model underwent a total of 1.5T tokens of pre-training across all phases. The final PLM met the project’s objectives, achieving strong performance in both English and Thai benchmarks, including Thai Exam evaluations.

\section{Pre-trained LLM Evaluation}

\subsection{Evaluation Benchmarks for Pre-trained Models}

To assess the model’s baseline intelligence, we adopted a logit-based evaluation method provided by im-eval-harness, focusing on multiple-choice formats. Specifically, this approach involves presenting the model with predefined answer options and measuring accuracy by evaluating the likelihood of selecting the correct choice. For further details, please refer to \cite{medium_llm_eval} and the im-eval-harness repository\footnote{\url{https://github.com/EleutherAI/lm-evaluation-harness}}.

We defined three key metrics to track model performance at this evaluation stage: Eng-H6, Thai-H6, and Thai-Exam.
\begin{itemize}
    \item \textbf{Eng-H6}: A consolidated benchmark suite widely used for LLM evaluation, integrating six datasets: MMLU~\cite{mmlu}, ARC~\cite{arc}, HellaSwag~\cite{hellaswag}, TruthfulQA~\cite{truthfulqa}, WinoGrande~\cite{winogrande}, and GSM8K~\cite{gsm8k}. Performance is measured by averaging individual scores across these datasets. This metric serves as a standard for evaluating foundational English-language knowledge.
    \item \textbf{Thai-H6}: A Thai-translated version of the Eng-H6 datasets. The six benchmarks were translated into Thai and refined through human verification to ensure accuracy.
    \item \textbf{Thai-Exam}: A Thai knowledge benchmarking dataset comprising multiple-choice questions sourced from Thai examinations\footnote{\url{https://huggingface.co/datasets/scb10x/thai_exam}}. This dataset was rapidly integrated into our project for evaluation purposes.
\end{itemize}

\subsection{Evaluation Results on Pre-trained Models}

\begin{table}[h!]
    \centering
    \begin{tabular}{c|l|c|c|c}
         \toprule
         & Models & Eng-H6 & Thai-H6 & Thai-Exam \\
         \midrule
         Our baseline & Phi-3-medium & 62.89 & - & 35.02 \\
         & GPT-3.5 (\textit{objective}) & 71.00 & 71.65 & 46.00 \\ 
         \midrule
         Thai LLMs & Typhoon-v1.5-72B & - & 64.80 & 61.70 \\
         & OpenThaiGPT-1.5-72B & - & 79.69 & - \\
         \midrule
         Ours & JTS-1-PLM & 78.95 & 71.80 & 62.32 \\
         \bottomrule
    \end{tabular}
    \caption{Performance comparison on Eng-H6, Thai-H6, and Thai-Exam. GPT-3.5 was the initial objective of this project and JAI-1-PLM achieves better performance than the objective.}
    \label{tab:plm_performance}
\end{table}

Table~\ref{tab:plm_performance} compares model performance across the Eng-H6, Thai-H6, and Thai-Exam benchmarks. The first key observation is the performance improvement achieved by upscaling and fine-tuning the phi-3-medium~\cite{phi3} model. While phi-3-medium initially showed weak results on Eng-H6 and Thai-Exam, the upgraded version demonstrated significant gains in both metrics. The second critical comparison is against GPT-3.5, which was our initial performance target across all three benchmarks. Our model achieved performance on par with or exceeding that of GPT-3.5 in these metrics. Additionally, we compared our model to Typhoon-v1.5-72B~\cite{typhoon}, a model released in May 2024 that was trained by adding Thai data to Qwen1.5-72B~\cite{qwen}. When evaluating Thai-specific metrics (Thai-H6 and Thai-Exam), JTS-1-PLM outperformed Typhoon-v1.5-72B. Finally, we conducted a comparison with OpenThaiGPT-1.5-72B, released in September 2024. This model appears to be built on Qwen2.5-72B~\cite{qwen2} with additional Thai data training. Due to the stronger foundation of the Qwen2.5-based architecture, OpenThaiGPT-1.5-72B achieved better Thai-H6 performance than JTS-1-PLM.
\section{Post-training: Supervised Finetuning}

The Supervised Finetuning (SFT) stage is crucial for enabling Pre-trained Language Models (PLMs), initially capable only of Next Token Prediction, to perform effectively in general-use conversational scenarios. High-quality training data is essential to achieve successful outcomes in this stage. Although open-source datasets for supervised finetuning have been made available, directly using these datasets presents two significant challenges: (1) many are low-quality synthetic data produced by earlier-generation models such as ChatGPT-3.5, and (2) they typically lack sufficient Thai language content.

To address these limitations, we employed a data augmentation strategy, using existing open-source data as seeds for regeneration, thereby enhancing data quality. Additionally, we developed specialized models—an English-to-Thai Translator and a Thai Fluency Reward model—to effectively extend English datasets into Thai. Moreover, we synthesized data specifically designed to improve understanding of Thai cultural nuances, further enriching our training set and resulting in a robust Thai-focused language model.

This session is structured as follows: (1) Data Generation Methods for each category, (2) Model Training Procedures, and (3) Analysis of Training Outcomes.

\subsection{Eng-Thai Translator and Thai Fluency Reward Model}

\subsubsection{Eng-Thai Translator}

To leverage extensive and diverse English datasets for SFT, we developed an English-to-Thai translation model. The foundational model was Upstage’s multilingual internal 10.7B LLM, which we finetuned for translation purposes. The training dataset comprised 0.5M English-Thai pairs from various sources, including large-scale open-source resources, news/media, government/legal texts, and web crawling.

This model outperformed GPT-3.5-turbo and showed comparable performance to GPT-4o-2024-05-13, achieving superior results on MMLU and ARC tasks but slightly lower performance on Flores. Despite its modest size and resource constraints, this translator significantly contributed to generating high-quality Thai data.

\begin{table}[h!]
    \centering
    \begin{tabular}{l|c|c|c}
    \toprule
        Model & FLORES & ARC & MMLU  \\
    \midrule
        GPT-4o-2024-05-13 & \textbf{46.5} & 62.7 & 72.2 \\
        GPT-4-0125-preview & 41.3 & 63.0 & 64.6 \\
        Solar-10.7B-translation-enth & 41.6 & \textbf{66.6} & \textbf{73.5} \\
    \bottomrule
    \end{tabular}
    \caption{English to Thai translation performance comparison. The measurements are BLUE scores on FLORES, ARC and MMLU.}
    \label{tab:my_label}
\end{table}

\subsubsection{Thai Fluency Reward Model}

To enhance the quality of synthetic Thai data, we developed a Thai Fluency Reward model. The foundational model was Solar Mini 1.2, a multilingual internal model from Upstage, which we finetuned using specially constructed fluency assessment datasets. The resulting reward model was then utilized for data validation.

Specifically, the reward model consisted of a 10.7B-sized LLM augmented with a linear layer capable of reward scoring. Fluency scores were generated using GPT-4o from diverse chat datasets, and pair-wise ranking data were constructed. Training was performed using the pair-wise ulytrainteract loss method, achieving approximately 90\% fluency classification accuracy.

We applied this model to evaluate and filter generated Thai data based on logit values, employing a systematic thresholding approach. The final threshold value selected for filtering was 0.88.

\subsection{Supervised Finetuning Data}

Our comprehensive SFT dataset comprises six primary categories: (1) General Chat Ability, (2) Instruction Following, (3) Wild Task Ability, (4) Thai Cultural Understanding, (5) Thai Safety, and (6) Long Context Understanding. Each category employed distinct data generation strategies, with Thai-language datasets undergoing additional reward-based filtering to ensure quality before their inclusion in training. 

\begin{table}[h!]
    \setlength{\tabcolsep}{2pt}
    \centering
    \begin{tabular}{l|c|c|c|c}
    \toprule
    Category & Language & Seed & \# of examples & \# of tokens \\
    \midrule
    1.General Chat  & EN, TH & Capybara & 65,664 & 70.0M \\
    2.Instruct Following  & EN, TH & Upstage's & 185,270 & 154.6M \\
    3.Task Ability   & EN, TH & flan\_v2, Upstage's  & 139,986 & 204.5M \\
    4.Thai Culture  & TH & Wikipedia & 169,276 & 22.5M \\
    5.Thai Safety   & TH & - & 5,235 & 0.5M \\
    6.Long Context  & EN & Upstage's & 13,369 & 158.4M \\
    \midrule
    \multicolumn{3}{r|}{Total} & 578,800 & 610.2M \\
    \bottomrule
    \end{tabular}
    \caption{SFT Data Statistics.}
    \label{tab:sft_data_statistics}
\end{table}

\subsubsection{General Chat Ability}

The simplest application of an LLM is general user chitchat. We utilized the widely adopted open-source Capybara dataset to construct this dataset. However, since Capybara only contained English data, we translated the data into Thai and refined it using OpenAI models to better align with Thai cultural contexts. Given the low complexity and broad accessibility of general knowledge, the data creation process remained relatively straightforward.

\subsubsection{Instruction Following}

Open-source datasets specifically for instruction following are limited. To overcome this, we leveraged the English instruction following dataset owned by Upstage, which was generated by synthesizing multi-turn instruction examples from an instruction pool using Upstage’s proprietary models. However, this dataset was exclusively English, lacking Thai-specific instruction following data, and suitable open-source LLMs for synthesis were unavailable.

To resolve this, we translated the instruction pool into Thai and synthesized new data using GPT-4o, incorporating a filtering logic to ensure appropriate Thai language generation.

\subsubsection{Wild Task Ability}

LLMs have broad potential applications. To expand the range of tasks LLMs can perform, we used datasets derived from Flan. However, these datasets lacked Thai data, limiting model performance. Directly utilizing Flan datasets proved suboptimal due to quality issues. Therefore, we refined the data by re-generating queries and answers using advanced open-source models, translating queries into Thai, and regenerating Thai answers accordingly.

Recognizing from MT-Bench evaluations that Coding, Math, Reasoning, and Extraction tasks required further improvement, we created additional specialized datasets based on queries from Upstage’s repositories. These datasets were translated and answers were generated using open-source models.

\subsubsection{Thai Cultural Understanding}

A Thai-focused LLM must extend beyond mere language proficiency to include an understanding of Thai cultural and historical contexts. We synthesized data using content from Thai Wikipedia, ensuring the model gained insights into Thai cultural nuances. Furthermore, we added specialized data about the Jasmine group to enhance cultural understanding.

\subsubsection{Thai Safety}

To manage sensitive information within Thai cultural contexts, we created queries involving sensitive topics and generated synthetic refusal responses, ensuring the model appropriately handles sensitive scenarios.

\subsubsection{Long Context Understanding}

Finally, to extend context handling capabilities, we developed long-context document-based QA data, thereby enhancing the model’s performance on RULER benchmarks.

\subsection{Supervised Finetuning Stages}

We conducted experiments using the assembled datasets to identify the optimal combination of training strategies. Preliminary experiments showed that training with individual datasets provided measurable improvements, but due to differences in task complexity and data diversity, it was necessary to determine an optimal training ratio in an integrated setting.

To address this, we designed multiple "recipes" that controlled the proportions of different dataset categories (Chat, Instruction Following (IF), Task, Culture, Safety, and Long-form data). Table \ref{tab:sft_recipe} summarizes two of the most effective recipes. For example, SFT.1 allocated 39.32\% of its training data to cultural datasets and 24.72\% to task-based data, whereas SFT.2 placed greater emphasis on IF data (32.01\%) while reducing the proportion of cultural data (29.25\%). The inclusion of safety-related examples was kept deliberately small in both cases (<2\%) to prevent overfitting while still instilling baseline alignment behavior. These variations allowed us to probe how shifts in balance between instruction-following and cultural data would affect downstream performance.

Benchmark evaluations of the resulting models are reported in Table \ref{tab:merge_performance}. Both recipes achieved competitive results across Thai benchmarks, but with differing strengths. SFT.1 achieved the best scores on IFEval-TH (77.00) and JAI-Hall-Bench (63.20), demonstrating strong reasoning and reduced hallucination. In contrast, SFT.2 achieved the highest score on MT-Bench-TH (7.45), suggesting better multi-turn conversational performance. To exploit the complementary strengths of the two recipes, we created a merged model by averaging their weights. The merged model produced a balanced outcome across all metrics: IFEval-TH (74.96), MT-Bench-TH (7.33), and JAI-Hall-Bench (62.10). Although not the single best on any individual benchmark, this merged model exhibited more stable and robust performance overall.

Through these experiments, we identified that both recipe design and post-hoc model merging are viable strategies for maximizing performance. The evidence suggests that carefully tuning dataset composition influences different capability axes (instruction-following, cultural grounding, and hallucination control). Moreover, merging independently trained models provided a practical method for reconciling trade-offs between these axes. After testing various ratios, we determined that a simple 1:1 weight merging of SFT.1 and SFT.2 was optimal, offering a well-rounded model suitable for deployment in diverse Thai NLP applications.

\begin{table}[h!]
    \centering
    \begin{tabular}{c|c|c|c|c|c|c}
        \toprule
        Recipe & Chat & IF & Task & Culture & Safety & Long \\
        \midrule
        SFT.1 & 10.15\% & 24.19\% & 24.72\% & 39.32\% & 1.62\% & 0.00\% \\
        SFT.2 & 11.34\% &  32.01\% & 24.19\% & 29.25\% & 0.90\% & 0.00\% \\
        \bottomrule
    \end{tabular}
    \caption{SFT data balance recipes. We evaluated more than 10 recipes and found that the two showed the best performance.}
    \label{tab:sft_recipe}
\end{table}

\begin{table}[h!]
    \centering
    \begin{tabular}{c|c|c|c}
    \toprule
        Model & IFEval-TH & MT-Bench-TH & JAI-Hall-Bench*  \\
    \midrule
        SFT.1 & \textbf{77.00} & 7.18 & \textbf{63.20} \\ 
        SFT.2 & 73.00 & \textbf{7.45} & 56.00 \\
        \midrule
        Merged & \underline{74.96} & \underline{7.33} & \underline{62.10} \\
    \bottomrule
    \end{tabular}
    \caption{Thai benchmark performance of SFT models. SFT.1 and SFT.2 are independently trained from the same pre-trained model, while Merged refers to a model obtained by averaging the weights of SFT.1 and SFT.2. Benchmark details are provided in Section 6.1. JAI-Hall-Bench* denotes an earlier version of the benchmark used for the final evaluation; it includes 298 samples excluding Wiki QA and focuses solely on hallucination detection without assessing fluency.}
    \label{tab:merge_performance}
\end{table}

\section{Post-training: Alignment-tuning}

This stage aims to incorporate human preferences into the model’s behavior. We employed Direct Preference Optimization (DPO) to perform alignment tuning and curated datasets across four critical dimensions: (1) chat alignment, (2) instruction-following alignment, (3) fluency alignment, and (4) hallucination alignment. These datasets were used to train the model to generate more preferred and reliable responses.

\subsection{Alignment-tuning Method}

Various methods have been proposed to align large language models (LLMs) with human preferences, beginning with Reinforcement Learning from Human Feedback (RLHF). In this work, we adopt Direct Preference Optimization (DPO), an offline reinforcement learning method, as an effective and computationally efficient approach for alignment tuning. DPO enables the model to learn directly from pairwise preference data without requiring an explicit reward model or costly online rollouts.

\subsection{Alignment-tuning Data}

The alignment tuning process was supported by four types of datasets, each tailored to improve a specific aspect of model alignment: chat alignment, Instruction Following Alignment, Fluency Alignment and Hallucination Alignment (Table \ref{tab:dpo_data_statistics}).

\begin{table}[h!]
    \centering
    \begin{tabular}{l|c|c|c}
    \toprule
    Category & Language & Seed & \# of examples \\
    \midrule
    1.General Chat  & TH & Capybara & 7,893 \\
    2.Instruct Following  & EN, TH & Upstage's & 67,462 \\
    3.Thai Hallucination   & TH & Wikipedia & 4,456 \\
    \midrule
    \multicolumn{3}{r|}{Total} & 79,811 \\
    \bottomrule
    \end{tabular}
    \caption{DPO Data Statistics. All examples have chosen and rejected response pairs.}
    \label{tab:dpo_data_statistics}
\end{table}

\subsubsection{General chat alignment}

To ensure that the model provides responses preferred by human users during open-ended conversations, we fine-tuned the model using pairwise preference data. Starting from a model trained via supervised fine-tuning (SFT), we generated five candidate responses per prompt. These responses were evaluated using GPT-4o (2024-11-20), and samples receiving high preference scores were labeled as positive while those with low scores were labeled as negative. This data was then used to train the model using DPO. The seed data for this process was derived from Thai-language data generated based on the Capybara dataset. A total of 7,893 pairwise samples were constructed.

\subsubsection{Instruction following alignment}

To enhance the model’s ability to follow user instructions in Thai, we analyzed failure cases encountered during data generation. From these, we curated positive and negative response pairs based on performance deviations. This instruction-following alignment dataset consists of 67,462 pairwise samples. 

\subsubsection{Hallucination alignment}

To address factual correctness, we used a Thai Wikipedia QA dataset and generated answers using the trained model. These outputs were evaluated by Claude to detect hallucinations. Responses judged to contain hallucinations were labeled as negative, while those without were labeled as positive. This process yielded a dataset specifically designed to reduce hallucinations during generation.

\section{Evaluation for Instruct Models}

\subsection{Evaluation Benchmarks for Instruct Models}

To define the optimization objectives for our model development, we established four key benchmarks to comprehensively evaluate the final performance of the Chat model. Each benchmark focuses on a distinct capability crucial to high-quality language modeling, including instruction-following, multi-turn dialogue, cultural fluency, and long-context understanding.

\begin{itemize}
    \item \textbf{IFEval-EN}: This benchmark assesses the instruction-following capabilities of Large Language Models. It is a benchmark focused on a set of verifiable instructions, such as specifying word counts or formatting requirements. The dataset is in English and is designed to evaluate chat or instruction-following fine-tuned language models.
    \item \textbf{IFEval-TH}: A Thai-translated and human-verified version of IFEval-EN. It measures how accurately LLMs follow verifiable instructions, with accuracy defined by whether the given instruction was correctly executed. This Thai version was created and released by the Typhoon project, which translated and validated the original IF-Eval samples.
    \item \textbf{MT-Bench-EN}: MT-Bench evaluates LLM performance in multi-turn dialogue scenarios across diverse domains. It consists of nine subcategories: Coding, Math, Reasoning, Extraction, Roleplay, Writing, Social Science, STEM, and Knowledge III. Model responses are assessed using the LLM-as-a-judge paradigm, with scores assigned on a 10-point scale.
    \item \textbf{MT-Bench-TH}: The Thai version of MT-Bench, which was translated and culturally adapted for the Thai language from the original MT-Bench-EN. We adopted the Thai version of MT-Bench-TH provided on the ThaiLLM leaderboard and used GPT-4o-2024-08-06 as the evaluation model, which was validated to provide stable results for benchmarking Typhoon2’s performance.
    \item \textbf{JAI-Hall-Bench}: Developed during the course of this project, this benchmark evaluates both cultural knowledge and Thai language fluency. It combines 298 project-specific samples and 1,000 QA items derived from Thai Wikipedia. Evaluation is based on two criteria: (1) the naturalness of the Thai-language response, and (2) the factual correctness of answers to Thailand-related questions. Both criteria are judged using the Claude-3.5-sonnet-20241022 model. A response is considered correct only if both aspects are rated as true. Final performance is reported as overall accuracy.
    \item \textbf{RULER}: RULER is a benchmark designed to evaluate long-context understanding in LLMs. It includes tasks such as retrieval, multi-hop tracing, and aggregation that go beyond simple information recall, requiring reasoning over extended input sequences. As context length increases, many models exhibit substantial performance degradation, highlighting the challenge of maintaining coherent reasoning in long-context settings.
\end{itemize}

\subsection{Evaluation Results on Instruct Models}

\begin{table}[h!]
    \setlength{\tabcolsep}{3pt}
    \centering
    \footnotesize
    \begin{tabular}{l|c|c|c|c|c|c}
        \toprule
         \textbf{Models} & 
         \begin{tabular}[c]{@{}c@{}}IFEval\\EN\end{tabular} & 
         \begin{tabular}[c]{@{}c@{}}IFEval\\TH\end{tabular} &  
         \begin{tabular}[c]{@{}c@{}}MT-Bench\\EN\end{tabular} & 
         \begin{tabular}[c]{@{}c@{}}MT-Bench\\TH\end{tabular} & 
         \begin{tabular}[c]{@{}c@{}}JAI-Hall\\Bench\end{tabular} & 
         RULER \\
         \midrule
         Typhoon2-70B~\cite{typhoon2} & \textbf{86.7} & 81.0 & \textbf{82.2} & 73.6 & 50.0 & 95.4 \\
         \midrule
         JAI-1-$\alpha$ (v1.2) & - & 69.0 & - & 64.0 & 21.0 & 92.0\\
         JAI-1 (v1.3) & \textbf{86.7} & \textbf{81.5} & 
         79.0 &
         \textbf{75.6} & \textbf{55.0} & \textbf{96.7}\\
         \bottomrule
    \end{tabular}
    \caption{Performance comparion between Typhoon2-70B and JAI-1. The initial version of JAI-1, JAI-1-$\alpha$, shows poor performance when compared from Typhoon2-70B. However, the latest version of JAI-1 provides superior performance on all target benchmarks.}
    \label{tab:performance}
\end{table}

Table~\ref{tab:performance} presents a performance comparison across the benchmarks defined in Section 8.1. The primary comparison target and final model benchmark is Typhoon2-70B~\cite{typhoon2}, a model released in December 2024 that builds upon the Llama3.3-70B architecture. Inheriting Llama’s foundational capabilities, Typhoon2-70B demonstrates strong performance on benchmarks like IFEval and MT-Bench. In our project, the earlier v1.2 model (JAI-1-$\alpha$) consistently underperformed compared to Typhoon2-70B across all defined benchmarks. However, the final v1.3 model (JAI-1) surpassed Typhoon2-70B in all evaluation metrics, achieving superior performance across the board.

\section{Conclusion}

JAI-1 is a Thai LLM with 75B parameters, developed by upscaling the phi-3-medium (14B) model, which has strong English baseline performance, through tokenizer adaptation, DUS, and MoE techniques, resulting in a 5x increase in model size. This approach expanded the parameter space by approximately 4x, enabling the storage of knowledge and information beyond English capabilities. By exploring optimized data recipes, we simultaneously improved performance in both English and Thai. Pre-training involved 1.5 trillion tokens, while post-training utilized over 600K examples. The pre-trained model achieved GPT-3.5-level performance on Eng-H6, Thai-H6, and Thai-Exam. The final JAI-1 model demonstrated superior performance compared to Typhoon2-70B on Thai-specific benchmarks like IFEval-TH, MT-Bench-TH, and JAI-Hall-Bench.

\bibliographystyle{unsrt}
\bibliography{reference}

\end{document}